\documentclass[runningheads]{llncs}
\usepackage{graphicx}
\usepackage{subfigure}
\usepackage{algorithm}
\usepackage{algorithmic}
\usepackage{makecell}
\usepackage{relsize}
\usepackage{amsmath,amsfonts,amssymb}
\usepackage{mathtools}
\usepackage{xfrac}
\usepackage{bm}
\usepackage{color}

\newcommand{\one}{\bm{1}}
\newcommand{\zero}{\bm{0}}

\DeclareMathOperator*{\argmin}{\arg\,\min}

\renewcommand{\L}{\mathcal{L}}
\newcommand{\x}{\bm{x}}
\newcommand{\y}{\bm{y}}

\newcommand{\U}{\bm{U}}
\newcommand{\I}{\bm{I}}

\newcommand{\w}{\bm{w}}

\newcommand{\m}{\bm{m}}

\newcommand{\loss}{\bm{\ell}}
\newcommand{\p}{\bm{p}}
\newcommand{\q}{\bm{q}}
\newcommand{\thet}{\bm{\theta}}
\newcommand{\B}{\mathcal{B}}
\newcommand{\R}{\mathbb{R}}
\newcommand{\X}{\mathcal{X}}

\newcommand{\RE}{\text{\tiny KL}}

\newcommand{\Red}  [1]{{\color{red}  {#1}}}

\begin{document}
%

\title{Exponentiated Gradient Reweighting for Robust Training Under Label Noise and Beyond}
\titlerunning{Exponentiated Gradient Reweighting for Robust Training}
%
\author{
Negin Majidi\inst{1}\and
Ehsan Amid\inst{2}\and
Hossein Talebi\inst{2}\and
Manfred K. Warmuth\inst{2}
}

%
\institute{
University of California, Santa Cruz, CA\\
\email{nemajidi@ucsc.edu}
\and
Google Research, Mountain View, CA\\
\email{\{eamid,\,htalebi,\,manfred\}@google.com}}


%
\maketitle              
\begin{abstract}
Many learning tasks in machine learning can be viewed as
taking a gradient step towards minimizing the average
loss of a batch of examples in each training iteration.
When noise is prevalent in the data, this uniform
treatment of examples can lead to overfitting to noisy
examples with larger loss values and result in poor
generalization. Inspired by the expert setting in on-line
learning, we present a flexible approach to learning
from noisy examples. Specifically, we treat each
training example as an expert and maintain a
distribution over all examples. We alternate between
updating the parameters of the model using gradient
descent and updating the example weights using the
exponentiated gradient update. Unlike other related methods, our approach handles a general class of loss functions and can be applied to a wide range of noise types and
applications. We show the efficacy of our approach for
multiple learning settings, namely noisy principal component analysis and a variety of noisy classification problems. 
\end{abstract}
\section{Introduction}

Machine learning (ML) models and deep neural networks have shown to be very powerful in learning from large amounts of data~\cite{he2016deep,devlin2018bert}. The majority of this advancement is attributed to supervised learning, typically leveraging a significant number of labeled training examples~\cite{ILSVRC15}. However, noisy examples can significantly deteriorate the generalization performance of the trained model~\cite{Zhu2004,ZhangBHRV16}. In some cases, the noise in the dataset can happen due to the noisy nature of data gathering process such as blurred images in satellite imagery~\cite{ansari2018noise}. In other cases, noise can occur due to human error or the intrinsic difficulty of the annotation task such as labeling medical images~\cite{liu2020deep}.

Many experiments have shown that corrupted data dramatically leads to poor generalization~\cite{algan2020label}. For instance, although deep neural networks show some level of robustness against label noise in classification tasks~\cite{rolnick2017deep}, they generally tend to overfit to the noisy examples, which negatively impacts the generalization performance of the model~\cite{liu2020early}. Alleviating this problem by improving the quality of the data gathering process is not always possible for several reasons such as the difficulty of the labeling task even for domain experts~\cite{liu2020deep}, disagreement among labelers with unknown levels of expertise~\cite{yan2014learning}, and the expensiveness of the annotation process~\cite{settles2008active}. Therefore, it is essential to account for the presence of input (i.e. feature) or label noise and develop methods to reduce their detrimental effects on ML models.

In this work, we apply the Weighted Majority (WM) method for 
aggregating the predictions of a set of ``experts''~\cite{littlestone1994weighted}. 
The WM method was initially developed for on-line learning:
In each round the algorithm must form its prediction by aggregating the predictions (also called the advice) from the set of experts. The underlying assumption is that there are a few experts (at least one) that are expected to perform well on the given task. Thus, the WM algorithm maintains a distribution over the experts and updates this distribution based on their performance in each round using the Exponentiated Gradient (EG) update~\cite{eg}. It has been shown that the WM algorithm leads to the optimal regret bounds in certain idealized settings when the loss is bounded~\cite{littlestone1994weighted,eg,hedge}.

In this work, we consider each example as an expert that contributes to guiding the ML model to improve its prediction. As in the WM algorithm, we maintain a distribution over the (mini-batch of) training examples. As the ML model is trained, the training  loss of the examples in each iteration is used to update the distribution over the examples using the EG update. The goal is to assign larger weights to cleaner (i.e. close to noise-free) examples while assigning smaller weights to the noisy ones. The underlying assumption for this approach is that the quality of an example is reflected in its training loss. For instance in a classification problem, noisy examples are expected to be misclassified and located far away from the classification boundary, thus incurring large loss values. Throughout the paper, we carefully analyze the validity of this assumption in each stage of training ML models and provide alternatives for cases where this assumption may not necessarily hold. 
Unlike most of the existing methods, our approach works for both supervised and unsupervised settings and we show experimentally that our method outperforms standard baselines for a variety of noise models. 

\subsection{Related work}

The problem of noisy data in ML has been thoroughly investigated in the literature~\cite{ghosh2017robust,zhang2018generalized,wang2019symmetric,li2020dividemix,co-teach+}. One approach is modifying the loss function to be robust against label noise; approaches like symmetric cross-entropy~\cite{wang2019symmetric}, robust cross entropy loss based on the Tsallis divergence~\cite{amid2019two}, bi-tempered generalization of the softmax with cross entropy loss~\cite{bi-tempered}, peer loss functions~\cite{Peerloss}, and Active-Passive loss~\cite{ma2020normalized}. A number of methods exploit the fact that ML models memorize noisy labels in the later phase of training and aim to prevent this behavior~\cite{liu2020early,liu2020earlylearning,xia2021robust}. Other approaches involve approximating the label noise transition matrix~\cite{JindalNC17,patrini2017making,sukhbaatar2015training,vahdat2017robustness}. Similar approaches to ours involve reweighting individual examples. In \cite{dataparams}, data parameters that represent the importance of the examples for a classification task are introduced to reduce the effect of noisy labels and updated using gradient descent (GD). Also,~\cite{wang2019derivative} proposes explicit weighting schemes on examples where weights are functions of probability of the labeled class. Meta-learning has also been investigated in some recent works to dynamically alter the loss functions to update the example weights based on the label noise~\cite{LearningToTeach,MentorNet,CurrNet,massiveNoisy}.

As for unsupervised learning, a number of approaches are proposed to improve the performance of the model when the data is noisy, specifically for PCA. Some existing methods utilize the robust norms as their loss functions to improve the robustness of the model~\cite{nie2014optimal,li2010l1,wang2014robust}. The RWL-AN algorithm~\cite{zhang2017auto} applies a reweighting of the examples using GD updates and eliminates the examples that incur large reconstruction errors. Using EG is inherently more suitable for non-negative weight updates than GD as the EG update implicitly imposes the non-negativity constraints and the projection to the probability simplex involves a simple division. EG also provides better guarantees in the expert setting than GD~\cite{eg}. To the best of our knowledge, our algorithm is the first applicable method for both supervised and unsupervised settings with noisy examples.

\section{Example Reweighting Using Exponentiated Gradient}
In this section, we discuss the core idea of example reweighting using the EG update. We first start with defining the weighted loss minimization problem and introduce EG Reweighting for the full-batch setting. Next, we develop a more general treatment for the mini-batch setting where only a subset of the examples is considered in a each step. Furthermore, we discuss the importance of regularizing the EG update to avoid ``over-punishing'' the noise-free but challenging examples. 

\subsection{EG Reweighting}
In many learning problems, the objective (i.e. the loss) function of the model consists of a sum over the individual losses of a given set of examples $\X = \{(\x_i, \y_i)\}_{i=1}^n$. Here, we denote by $\x_i$ and $\y_i$ the input instance and the target label of the $i$-th example, respectively. The optimization problem is then to minimize the overall loss w.r.t. the model parameters $\thet \in \R^m$,
\begin{equation}
    \label{eq:loss_func}
    \thet^* = \argmin_{\thet\in\R^m} \L(\thet|\mathcal{X}) \text{\,\,\, where \,\,\,} \L(\thet|\mathcal{X})  = \frac{1}{n} \sum_{i=1}^{n} \ell(\thet\vert \x_{i},\y_{i})
 \end{equation}
 Instead of directly minimizing the objective in Eq.~\eqref{eq:loss_func}, it is common to maintain an estimate of the solution and proceed by iteratively taking a gradient step to update the current model parameters. In its simplest form, the gradient descent (GD) update at iteration $t$ consists of a small step in the direction of the steepest descent direction of the total loss at $\thet^t$
 \begin{equation}
 \label{eq:gd}
     \thet^{t+1} = \thet^{t} - \eta_{\theta}\, \frac{1}{n}\sum_i \nabla \ell(\thet^t|\x_i, \y_i)\, ,
 \end{equation}
 where $\eta_{\theta} > 0$ is the learning rate. This can alternatively be viewed as the solution of the following regularized loss minimization problem\vspace{-0.0cm} 
 \[
 \thet^{t+1} = \argmin_{\thet\in\R^m}\,\big\{{\underbrace{\vphantom{\sfrac{1}{n}\,\sum_i \widehat{\ell}(\thet|\x_i, \y_i, \thet^t)}\sfrac{1}{2\eta_{\theta}}\,\Vert \thet - \thet^t\Vert^2}_{\mathclap{\text{regularizer}}}} 
    +  { \underbrace{\sfrac{1}{n}\,\sum_i \widehat{\ell}(\thet|\x_i, \y_i, \thet^t)}_{\mathclap{\text{(linearized) loss}}}}\big\}\,,\]
    
\vspace{-0.1cm}
 \noindent where $\widehat{\ell}(\thet|\x_i, \y_i, \thet^t) \coloneqq \ell(\thet^t|\x_i, \y_i) + \nabla \ell(\thet^t|\x_i, \y_i) \cdot (\thet - \thet^t)$ is a linearized version of the loss using the first order Taylor approximation at $\thet^t$. The goal of the $\mathrm{L}_2$-regularizer used to motivate the GD update is to prevent the updated parameters from drifting away from the current estimate $\thet^t$. This regularized view of the step will be crucial for developing the EG update.
 
The sum in Eq.~\eqref{eq:loss_func} implies that each example $(\x_i, \y_i) \in \X$ contributes equally to the total loss of the model. In other words, Eq.~\eqref{eq:loss_func} can be viewed as a weighted sum over the loss of the examples where the weights of all examples are identical and equal to $\sfrac{1}{n}$. This uniform treatment of the examples may result in overfitting to noisy examples which incur larger loss values than the noise-free (i.e. clean) examples (see Fig.~\ref{fig:fashion-blur}(b)). Instead, we consider maintaining a more general distribution over the examples. Our goal is to assign smaller weights to the noisy examples compared to the clean examples. Thus, the new objective becomes to jointly minimize the weighted total loss w.r.t the model parameters $\thet\in\R^m$ as well as the example weight vector
$\w$ that lies on the probability $n$-simplex $\Delta^{n-1} = \{\p\in \R^n\vert\, \p \geq \zero\,\, \&\,\, \sum_i p_i = 1\}$: 
\begin{equation}
    \label{eq:cost_func_w}
    \{\thet^*, \w^*\} = \!\!\!\!\argmin_{\{\thet \in \R^m,\, \w \in \Delta^{n-1}\}}\!\! \L(\thet, \w|\mathcal{X}) \text{\, where \,} \L(\thet, \w |\mathcal{X})  = \sum_{i=1}^{n} w_{i} \ell(\thet\vert \x_{i},\y_{i})\, ,
\end{equation}
Note that uniform weights, i.e. $\w = \sfrac{1}{n}\, \one$ recovers Eq.~\eqref{eq:loss_func} as a special case.

For a fixed set of weights $\w$, the $\thet$ parameter update proceeds similarly to the previous case, which results in the following update
 \begin{equation}
 \label{eq:weighted-gd}
     \thet^{t+1} = \thet^{t} - \eta_{\theta}\, \sum_i w_i \nabla \ell(\thet^t|\x_i, \y_i)\, .
 \end{equation}
This corresponds to a weighted gradient descent step. On the other hand, a na\"ive minimization of~Eq~\eqref{eq:cost_func_w} w.r.t the example weights $\w$ (while keeping $\thet$ fixed) assigns a unit weight to the example with the smallest loss value and zero weights to the remaining examples. Instead, we consider a similar iterative procedure which minimizes a regularized version of the objective.  The exponentiated gradient (EG) update~\cite{eg}  is motivated by minimizing the linear loss $\w \cdot \loss^t$ w.r.t. $\w$ along with a relative entropy (a.k.a. Kullback-Leiber) divergence as the regularizer, 
\begin{equation}
    \label{eq:eg-obj}
    \w^{t+1} = \argmin_{\w \in \Delta^{n-1}}\,\big\{\underbrace{\sfrac{1}{\eta_w}\,D_{\RE}(\w, \w^t)}_{\text{regularizer}} + \underbrace{\vphantom{\sfrac{1}{\eta}\,D_{\RE}(\w, \w^t)}\w \cdot \loss^t}_{\text{linear loss}}\big\}\, ,
\end{equation}
where again $\eta_w > 0$ is a positive learning rate and $[\loss^t]_i = \ell(\thet^t\vert\x_i, \y_i)$ denotes the vector of loss values evaluated at $\thet^t$. Also,
\[
D_{\RE}(\p, \q) \coloneqq \sum_{i=1}^n p_i \log\frac{p_i}{q_i} - p_i + q_i\, ,
\]
is the KL divergence between probability vectors $\p, \q \in \Delta^{n-1}$. Minimizing~\eqref{eq:eg-obj} by adding a Lagrange multiplier to enforce the constraint $\sum_i w_i = 1$ yields the EG update~\cite{yan2014learning} (Note that the divergence enforces the $w_i\ge 0$ constraints):
\begin{equation*}
    \label{eq:eg-update}
    \w^{t+1} = \frac{1}{W^t}\, \w^t \, \exp(-\eta\, \loss^t)\, ,
\text{ where $W^t \coloneqq \sum_i w^t_i\, \exp(-\eta\,\ell^t_i)$. }
    \tag{EG Update}
\end{equation*}
The EG update can equivalently be viewed as an Unnormalized Exponentiated Gradient (EGU) update~\cite{eg} on the weights, followed by a Bregman projection~\cite{bregman} w.r.t. the KL divergence onto the probability simplex $\Delta^{n-1}$, that is,
\begin{align*}
    \widetilde{\w}^{t+1} & = \argmin_{\w \ge \zero} \,\big\{\sfrac{1}{\eta}\,D_{\RE}(\w, \w^t) + \w \cdot \loss^t\big\}\\
    & = \w^t \, \exp(-\eta\, \loss^t) \tag{EG Unnormalized}\, ,\\
    \w^{t+1} & = \argmin_{\w\in \Delta^{n-1}} D_{\RE}(\w, \widetilde{\w}^{t+1}) = \frac{1}{W^t}\, \widetilde{\w}^{t+1}\, .\tag{Projection}
\end{align*}
In this case, the Bregman projection simply corresponds to dividing by the sum of the unnormalized weights $W^t$. This view of the updates allows us to generalize the formulation to the mini-batch setting (where only a subset of examples is considered in the update), as shown in the next section. Note that the multiplicative form of the update along with the normalization ensures that $\w^{t+1} \in \Delta^{n-1}$.

\subsection{Mini-batch Setting}
A common practice for training is to use a mini-batch of examples at each step instead using the full-batch of examples. That is, a subset of indices $\B^t \subseteq [n]$ corresponding to the subset of examples $\X^t \subseteq \X$ is used at iteration $t$. Thus, we consider a more general approach that includes the full-batch setting as a special case. Specifically, instead of maintaining a normalized weight $\w^t \in \Delta^{n-1}$, we can consider the unnormalized weights $\widetilde{\w}^t \in \R^n_+$ throughout the training and apply the EGU weight updates on the coordinates (i.e. weights) that are present in the mini-batch $\X^t$,
\begin{equation}
\label{eq:eg-batch}
\widetilde{w}^{t+1}_i = \widetilde{w}_i^t\, \exp(-\eta_w\, \loss(\thet^t|\, \x_i, \y_i))\quad \text{ for \,\,} i \in \B^t\, .
\end{equation}
The EGU update is then followed by a projection onto the simplex to form the following weighted loss for updating the model parameters $\thet \in \R^m$,
\begin{equation}
\label{eq:obj-main-mini}
\min_{\thet\in\R^m} \Big\{\frac{1}{W_{\B^t}}\,\sum_{i\in\B^t} \widetilde{w}^{t+1}_i\,\ell(\thet\vert\x_i, \y_i)\Big\}\, ,
\text{ where $W_{\B^t} \coloneqq \sum_{i\in \B^t} w^{t+1}_i$.}
\end{equation}

The EG weight update is followed by a parameter update using the gradient of the objective w.r.t. $\thet$ and performing a gradient descent step as before,
\[
\thet^{t+1} = \thet^{t} - \eta_\theta\, \frac{1}{W_{\B^t}}\sum_{i\in\B^t} \widetilde{w}^{t+1}_i \nabla \ell(\thet^t\vert\x_i,\y_i)\, ,
\]
where $\eta_\theta > 0$ denotes the parameter learning rate. Note that now the sample gradient $\nabla \ell(\thet^t\vert\x_i,\y_i)$ is scaled by the factor 
$\frac{\widetilde{w}^{t+1}_i}{W_{\B^t}}$.\footnote{Our mini-batch updates are closely related to the specialist approach developed for on-line learning~\cite{specialist}, which would use the (normalized) EG to update the example weights of the current mini-batch, always renormalizing so that total of weight of the mini-batch examples remains unchanged. Our method of maintaining unnormalized weights combined with GD updates on the reweighted gradients performs better experimentally.}

\subsection{Regularized Updates}
The main goal of example reweighting is to reduce the effect of noisy examples by penalizing the example with large loss values. However, this reweighting procedure also affects the harder examples which may initially have large loss values, but are eventually learned by the model later in the training. This side effect is desirable initially to allow the model to learn from the easier examples to form a better hypothesis. A similar approach is considered in curriculum learning where the level of difficulty of the training examples is increased gradually over the course of training~\cite{curriculum}. However, in the case of EG updates, there is no mechanism to recover an example which incurs large loss values earlier in the training. In this section, we propose a regularized version of the updates that alleviates this problem.

Consider the following modified formulation of the EG update in Eq.~\eqref{eq:eg-obj},
\begin{equation}
\label{eq:reg-eg-obj}
\w^{t+1} = \argmin_{\w \in \Delta^{n-1}}\Big\{\sfrac{1}{\eta}\, D_{\RE}(\w, \w^t) + \w \cdot \loss^t \!\!\underbrace{-\,\, \sfrac{1}{\gamma}\, H(\w)}_{\text{maximize entropy}}\Big\}\, ,
\end{equation}
where the last term $H(\w) \coloneqq -\sum_i \big(w_i \log w_i - w_i\big)$ is the entropy of the distribution $\w$ and $\gamma > \eta$ is a regularization constant. The effect of adding the negative entropy of $\w$ to the objective function is to bring the weights closer towards a uniform distribution after the update. By introducing a Lagrangian multiplier $\lambda$ into Eq.~\eqref{eq:reg-eg-obj} to enforce the constraint $\sum_i^{n} w_i = 1$, we have 
\begin{equation}
    \label{eq:reg-eg}
    \w^{t+1} = \argmin_{\w \in \R^n}\,\big\{\sfrac{1}{\eta}\,D_{\RE}(\w, \w^t) + \w \cdot \loss^t - \sfrac{1}{\gamma}\, H(\w) + \lambda (\sum_{i=1}^{n} w_i - 1)\}\, ,
\end{equation}

Setting the derivatives to zero yields
\begin{equation}
    \label{eq:eg-reg2}
    (\sfrac{1}{\eta} + \sfrac{1}{\gamma} )\, \log \w - \sfrac{1}{\eta} \log \w^t +  \loss^{t} + \lambda = 0
\end{equation}

Let $ \frac{1}{\eta'} := \frac{1}{\eta} + \frac{1}{\gamma}$. For $\eta > 0$ and $\gamma > 0$, we have $\eta' \leq \eta$. Let $0 \leq r := \frac{\eta}{\eta'} \leq 1$. Refactoring Eq.~\eqref{eq:eg-reg2} in term of $r$ and enforcing the constraint yields,
\begin{equation*}
\w^{t+1} = \frac{1}{W^t}\, \big(\w^t \, \exp(-\eta\, \loss^t)\big)^r\, \,\,\text{ for \,\,} 0 \leq r \leq 1\,, \tag{Regularized EG Update}
\end{equation*}
where $W^t \coloneqq \sum_i \big(w^t_i\, \exp(-\eta\,\ell^t_i)\big)^r$. Note that $r=1$ recovers the vanilla EG update whereas $r=0$ sets the updated weights to a uniform distribution, i.e. $\w^{t+1} = \sfrac{1}{n}\,\one$. Model training using regularized EG update for example reweighting is summarized in Algorithm~\ref{alg:eg}.

\begin{figure*}[t]
{\centering
\begin{algorithm}[H]
    \centering
    \caption{Example Reweighting Using the Exponentiated Gradient Update}\label{alg:eg}
    \begin{algorithmic}
        \STATE \textbf{input:} training examples $\X$ of size $n$, model parameters $\thet$, model parameters and weights learning rates $(\eta_\theta, \eta_w)$, regularizer factor $0 \leq r \leq 1$\smallskip
        \STATE initialize $\widetilde{\w}^0 = \one_n$
        \FOR{$t=0$ to $T-1$}
        \STATE for a given data batch $\mathcal{X}^t \subseteq \X$ indexed by $\B^t \subseteq [n]$
        do\smallskip
        \STATE $\bullet$ update the weights:\,\,\, $\widetilde{w}^{t+1}_i = \big(\widetilde{w}^{t}_i\, \exp(-\eta_w\,\ell(\thet^t|\x_i, \y_i))\big)^r$ for $i \in \B^t$\smallskip
        \STATE $\bullet$ calculate the sum:\,\,\, $W^t_{\B} = \sum_{i \in \B^t} \widetilde{w}^{t+1}_i$\smallskip
        \STATE $\bullet$ update the model parameters:\,\,\, $\thet^{t+1} = \thet^{t} - \eta_\theta\, \frac{1}{W_{\B^t}}\sum_i \widetilde{w}^{t+1}_i \nabla \ell(\thet^t\vert\x_i,\y_i)$
        \ENDFOR
    \end{algorithmic}
\end{algorithm}
}
\vspace{-1cm}
\end{figure*}



\section{Capturing Noisy Examples in Model Training}

In this section, we discuss the practical considerations for applying EG Reweighting for noise-robust training. First, we discuss the problem of overfitting to noise and develop a learning rate schedule for the EG update that we found beneficial in practice. We also show the effect of the regularization on the updates. Next, we discuss the limitations of the EG Reweighting approach using the training loss for the weight updates, especially when applied to deep neural networks. We provide practical solutions in this direction and validate our approach in the experimental section. As our motivating example, we consider a simple convolutional neural network (two convolutional layers of size $32$ and $64$, followed by two dense layers of size $1024$ and $10$) trained on different noisy versions of the Fashion MNIST dataset~\cite{fashion}. We use a SGD with heavy-ball momentum ($0.9$) optimizer with a fixed learning rate of $0.1$ and train the model for $80$ epochs with a batch size of $1000$. The baseline model achieves $91.35\pm0.38\%$ test accuracy.

\subsection{When to Apply the EG Updates?}
The update in Eq.~\eqref{eq:eg-batch} adjusts the weight of an example based on the value of the training loss of the example. This update entails the assumption that the noise may move the examples far away from the decision boundary, thus causing the example to induce larger loss values. We will discuss the validity of such assumption more thoroughly for different cases in the next section. The main question that we aim to address in this section is at which stage of training the model, the loss of the example may be indicative of noise?

\begin{figure*}[t!]
\vspace{-0.25cm}
\begin{center}
    \subfigure[]{\includegraphics[width=0.24\linewidth]{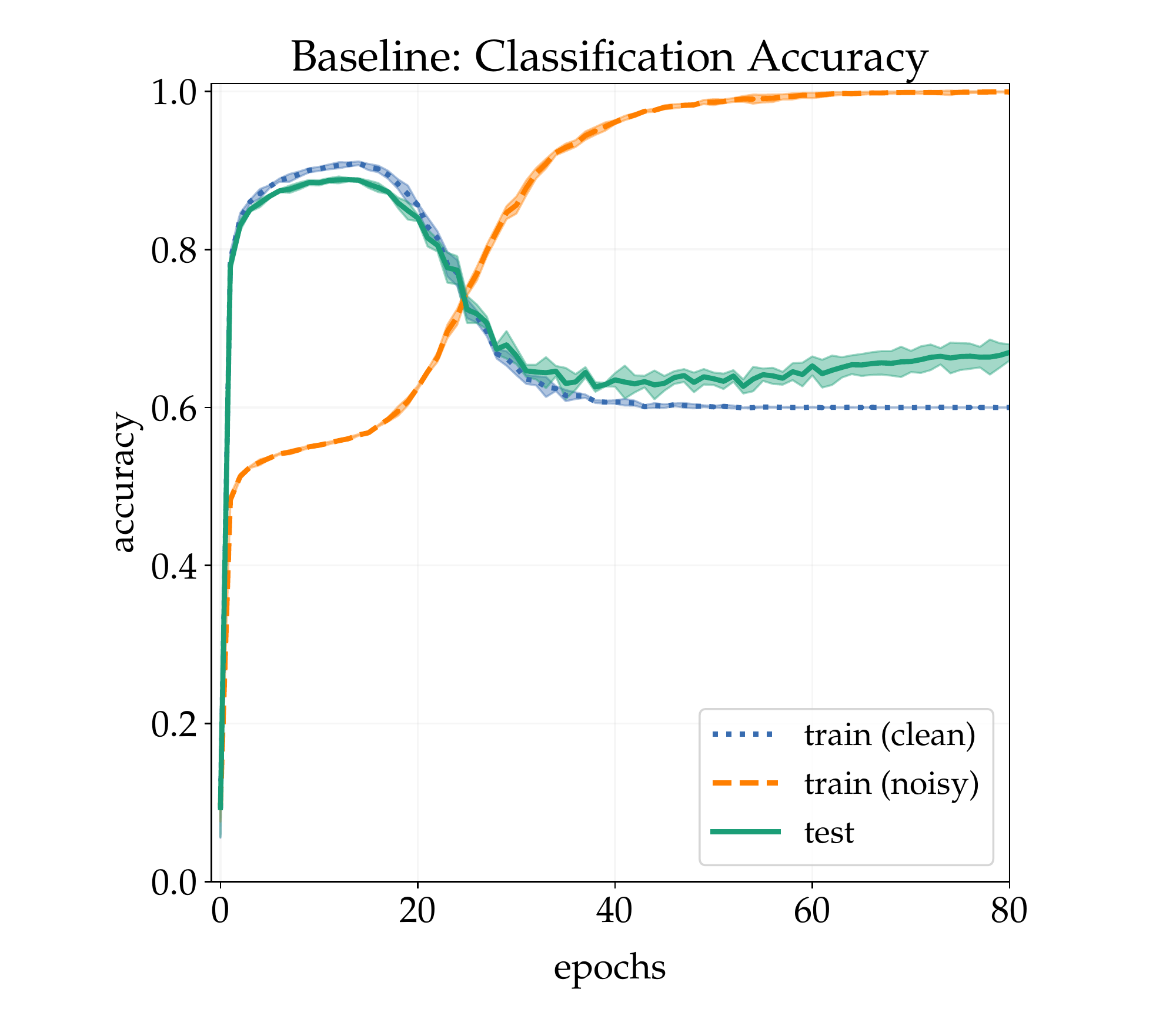}}  \subfigure[]{\includegraphics[width=0.24\linewidth]{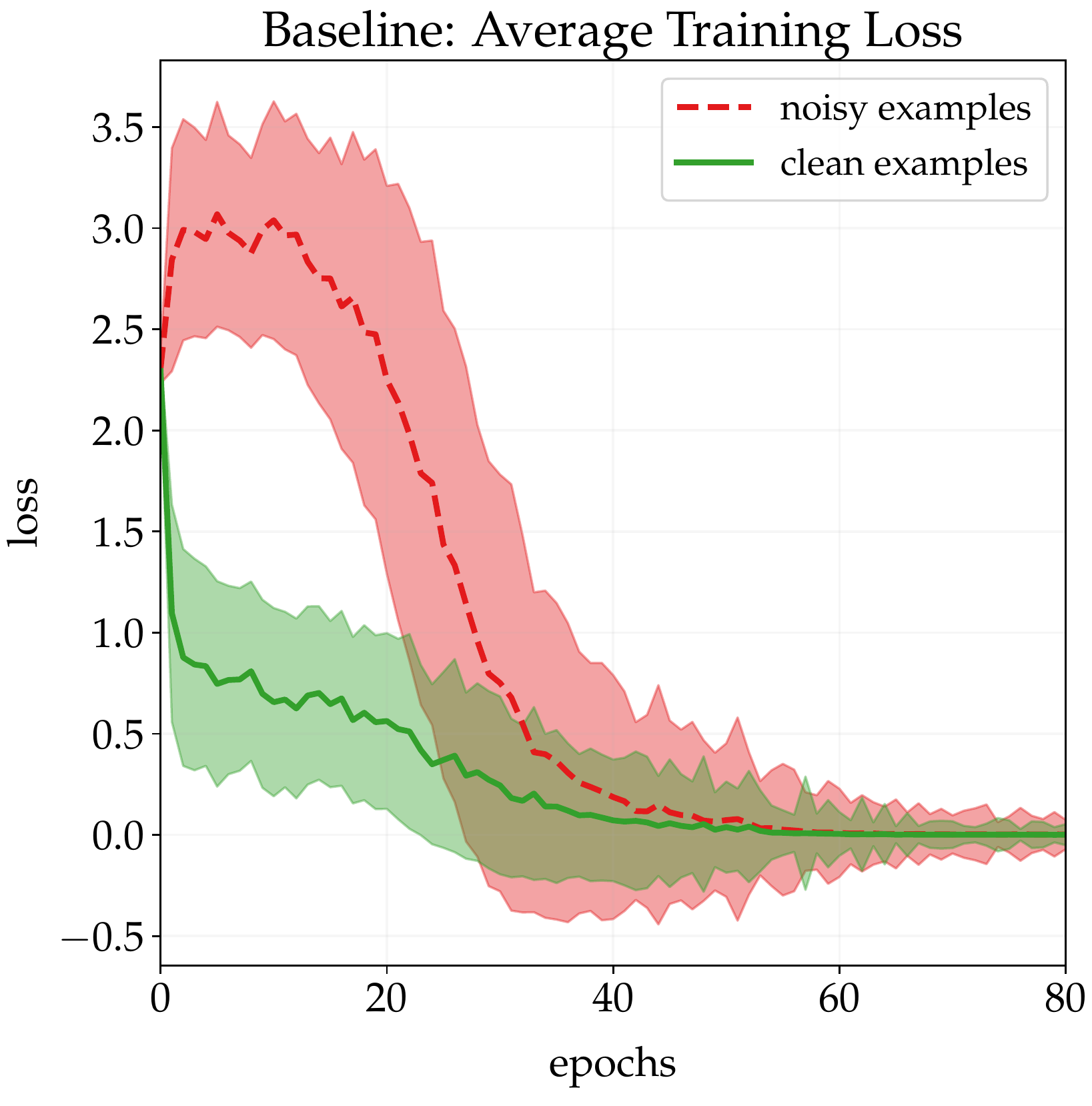}}
    \subfigure[]{\includegraphics[width=0.24\linewidth]{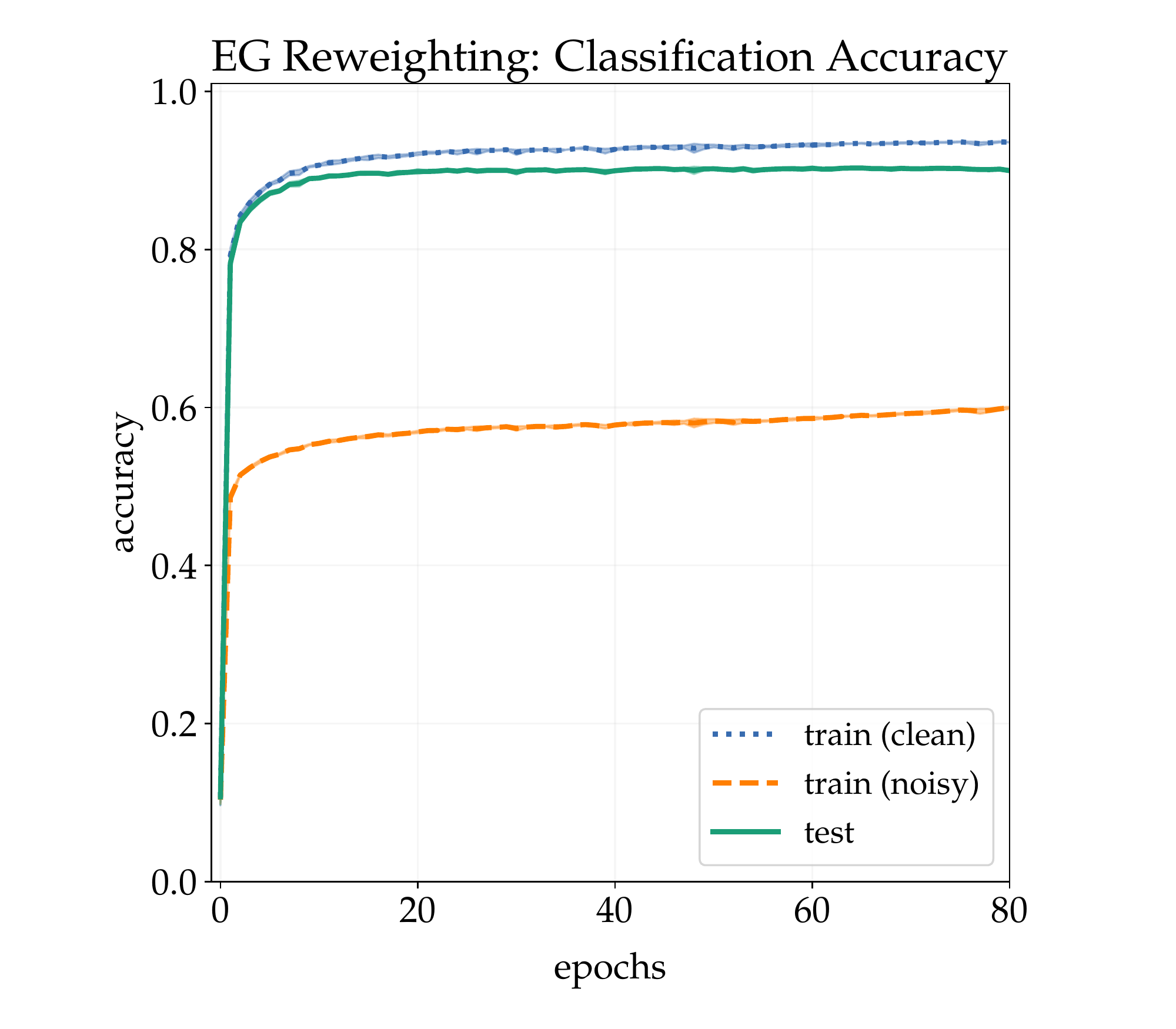}}  \subfigure[]{\includegraphics[width=0.24\linewidth]{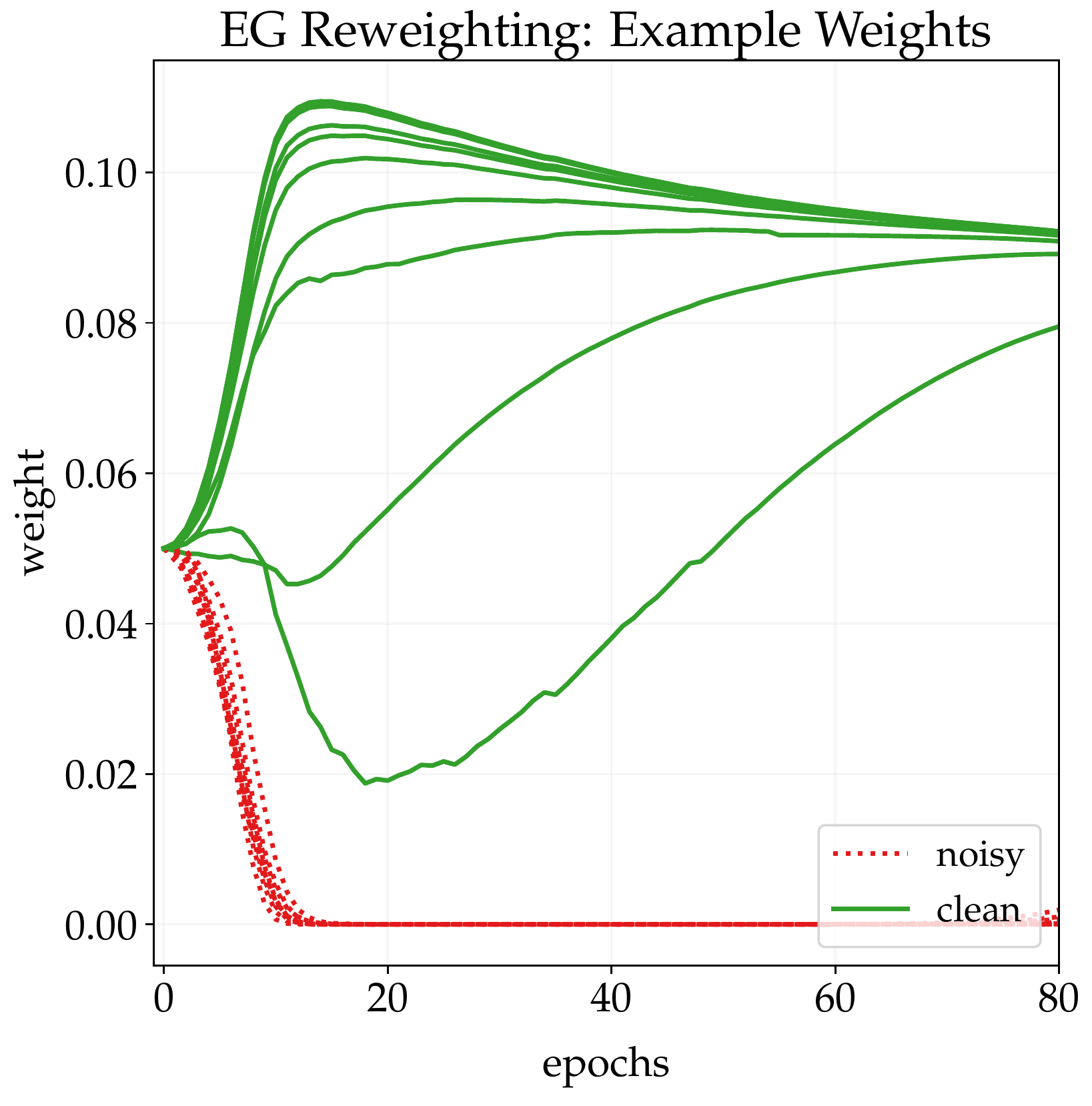}}\\[-1mm]

    \subfigure[]{\includegraphics[width=0.24\linewidth]{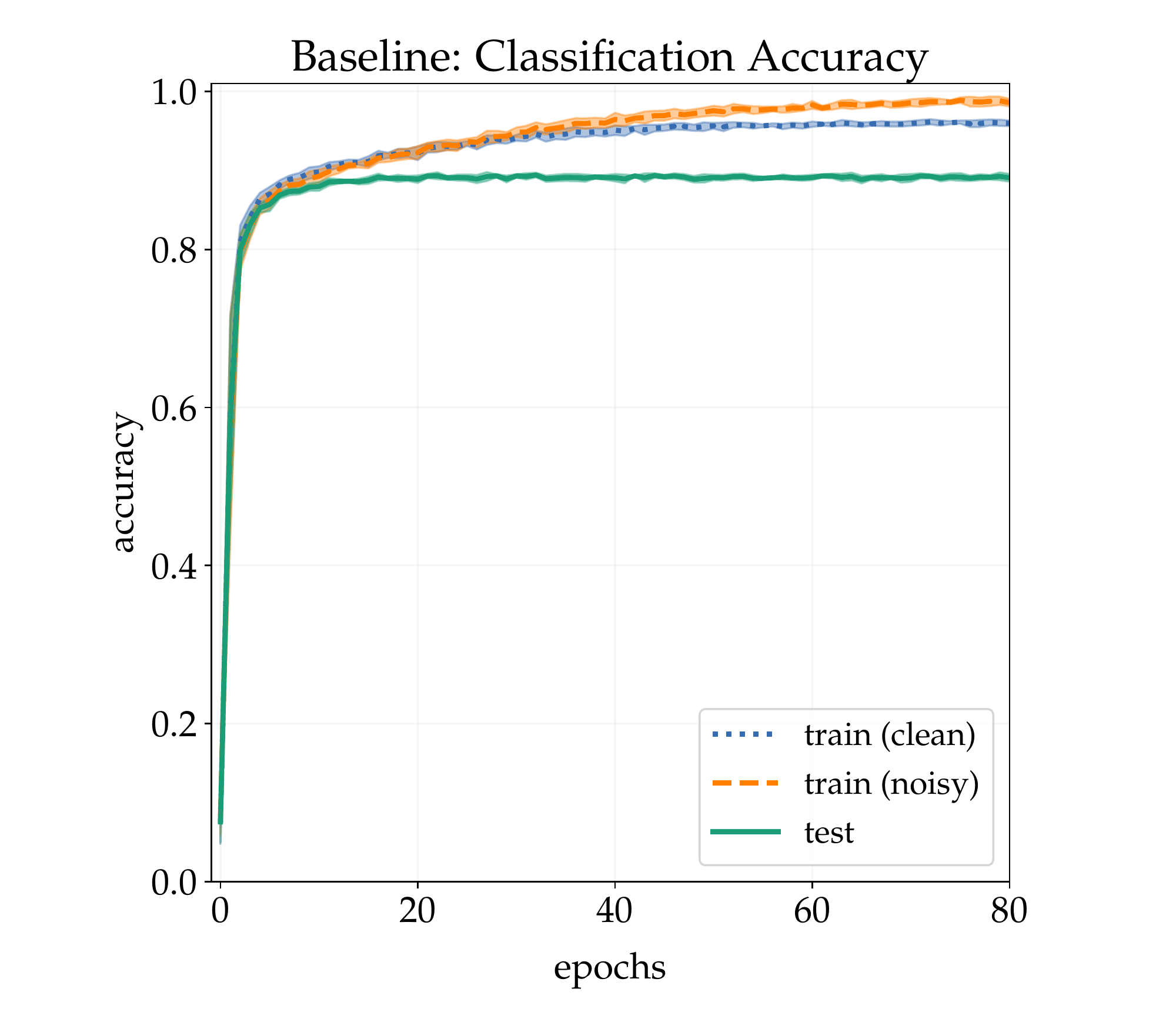}}  \subfigure[]{\includegraphics[width=0.24\linewidth]{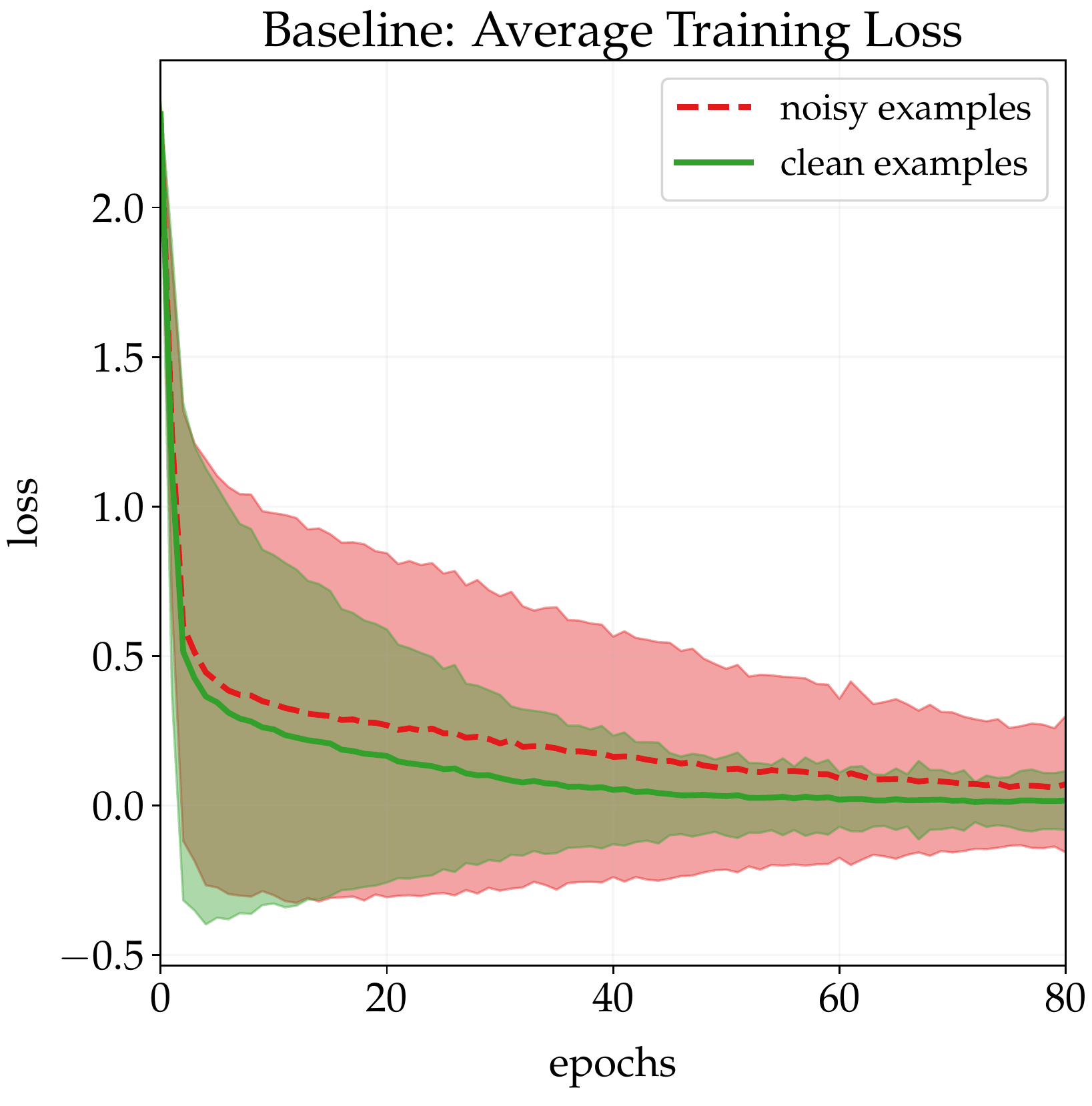}}
    \subfigure[]{\includegraphics[width=0.24\linewidth]{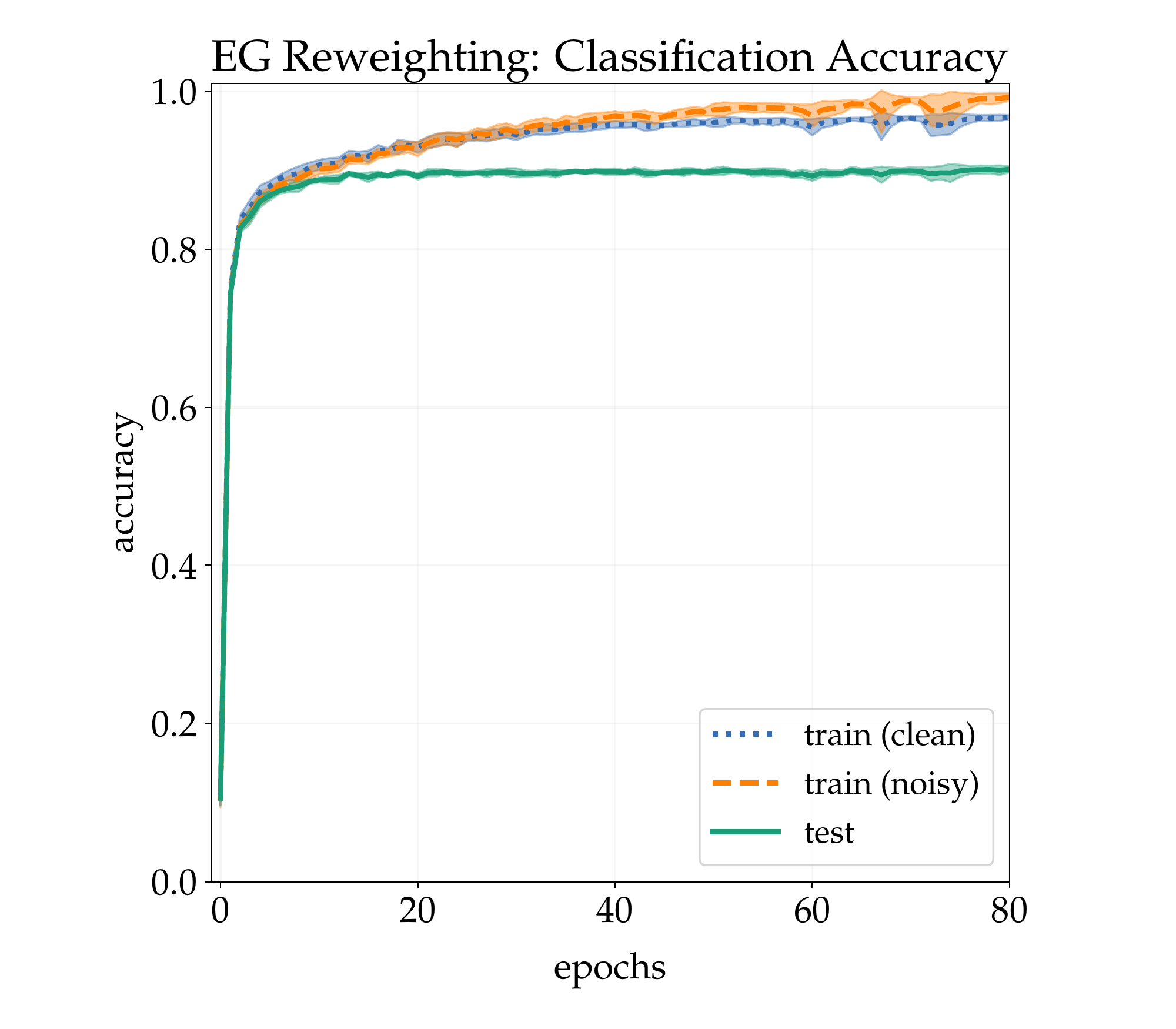}}  \subfigure[]{\includegraphics[width=0.24\linewidth]{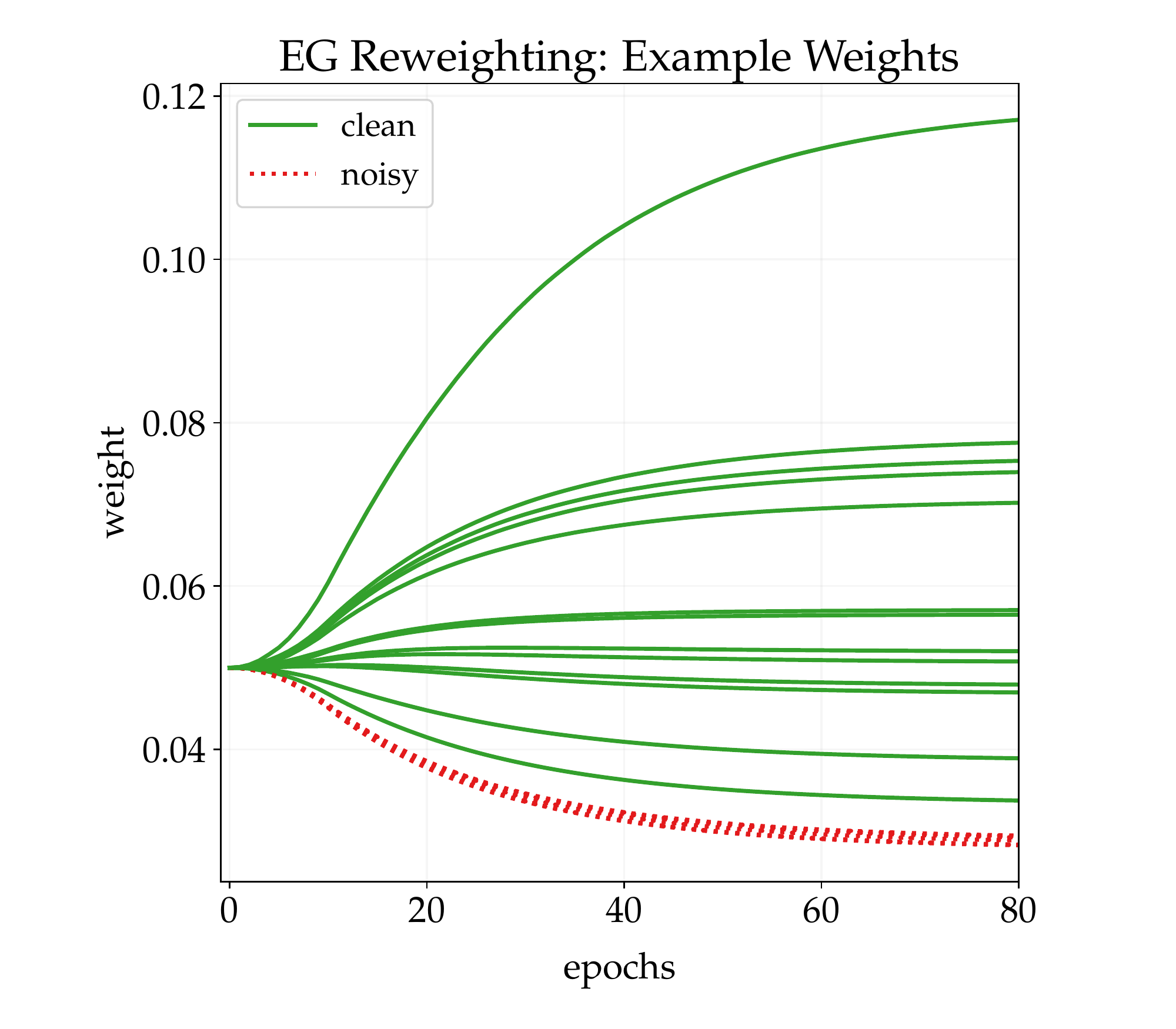}}
    \vspace{-0.25cm}
    \caption{\textbf{Fashion MNIST Classification -- Label noise :} (a) The model overfits to noisy examples after $\sim\!15$ epochs. (b) noisy examples incur relatively larger training loss values compared to the clean examples. (c) EG Reweighting successfully captures the noise and improves generalization. (d) A normalized subset of $20$ weights shows how clean but hard to classify examples recover later on via the EG regularization. \textbf{Blur noise:} (e) The (clean) train and test accuracy does not reveal any major overfitting. (f) Average training loss values for the clean and noisy examples are roughly the same. (g) We apply EG Reweighting on a pseudo-loss, consisting of the negative variance of the Laplacian operator~\cite{laplacian} applied to each image. EG Reweighting successfully improves the performance. (h) A normalized subset of $20$ weights which indicates the progression of the weights for clean and noisy examples.}
    \label{fig:fashion-blur}
    \end{center}
    \vspace{-0.5cm}
\end{figure*}

In general, model training with noise can be split into two phases: i) initial warm-up and ii) overfitting to noise~\cite{xia2021robust,liu2020earlylearning}. In the early phase of training, the model starts from an initial solution and gradually forms a hypothesis based on the given set of noisy examples. In this phase, the model parameters are still close to the initial solution, thus the model behaviour is mainly dominated by a regularization effect similar to early stopping. At this point, the solution may not be highly descriptive of the (clean) data, but at the same time has not overfit to the noisy examples. This is shown in Fig.~\ref{fig:fashion-blur}(a) where we train the model on the Fashion MNIST dataset with $40\%$ symmetric label noise. The classification accuracy on both the clean train set (not available to the model) as well the held-out test set improves up to around $15$ epochs. However at this stage, the large loss (thus, the gradient) of noisy examples starts to dominate the model training (Fig.~\ref{fig:fashion-blur}(b)). Thus, the model starts to overfit to the noise and thus, sacrificing generalization. 

Based on this observation, the effect of example reweighting should gradually increase during the warm-up phase and then decrease over time once the overfitting is prevented. Thus, we propose using the following learning rate schedule for the EG Reweighting procedure, which we found effective in our experiments: \emph{a linear warm-up in the early stage followed by a decay over time.} In this example, we apply a linear warm-up, up to $0.1$, in the first $20$ epochs followed by an exponential decay by a factor of $0.95$ per epoch. We use $r=0.98$. Fig.~\ref{fig:fashion-blur}(c)  shows that EG Reweighting with the proposed schedule successfully captures the noise and improves generalization. Fig.~\ref{fig:fashion-blur}(d)  shows a subset of $20$ weights (normalized by their sum). As can be seen, the weights for some of the clean examples drop initially. These clean examples are inherently harder for the model to classify initially, thus incurring comparatively larger training losses. However, the regularization in the EG updates allows these examples to recover later on.

\subsection{When is the Noise Reflected in the Training Loss?}
The underlying assumption for example reweighting in Algorithm~\ref{alg:eg} is that the noisy examples (after the initial warm-up stage) incur much higher loss values than the clean examples. Here, we show a case where such an assumption does not hold. For this, we consider the same model as in the previous section for classifying the Fashion MNIST dataset~\cite{fmnist}. However, instead of adding label noise, we consider corrupting a subset of examples by adding blur noise to the input image. Specifically, we sample $40\%$ of the images in the training set randomly and convolve these images with a Gaussian filter with a radius of $2$. As can be seen in Fig.~\ref{fig:fashion-blur}(a), although the generalization performance of the model degrades compared to the noise-free baseline model, no clear overfitting manifests during training and the noisy and clean examples incur roughly the same average training loss values throughout (Fig.~\ref{fig:fashion-blur}(b)). Thus, any method, including vanilla EG Reweighting, which aims to reduce the effect of noisy examples based on the training loss may fail to improve the performance. 

In such settings, having access to a method that can reflect the noise in the data can be highly beneficial. Consequently, we can leverage this information for reweighting the examples using the EG update. In particular, we consider the signal from such a model as a \emph{pseudo-loss} for EG, replacing the training loss in the weight update in Algorithm~\ref{alg:eg}. In the case of the Gaussian blur in our toy example, we calculate the variance of the Laplacian operator~\cite{laplacian}. A lower value of variance indicates smooth edges, thus suggesting a higher rate of blurring. Therefore, we consider the negative of the variance of the Laplacian operator as the pseudo-loss. We use the same hyperparameters and learning rate schedule as the previous section, but set $r=1$. Fig.~\ref{fig:fashion-blur}(c) shows that using this pseudo-loss improves the generalization performance of the model (from $89.07\pm0.50$\% test accuracy to $90.13\pm0.38$\% test accuracy). Also, Fig.~\ref{fig:fashion-blur}(d) illustrates a subset of $20$ weights (normalized). As the final remark, note that in many cases the pseudo-loss does not depend on the model parameters, thus does not vary during training. In such cases (as is in this example), the pseudo-loss needs to be calculated only once for each example (when visited for the first time). We will consider a more realistic setup in the next section.

\section{Experiments}

In this section we evaluate the performance of our EG Reweighting algorithm through extensive unsupervised and supervised learning experiments.  Our first set of experiments involve the unsupervised  problem of learning the principal component, i.e. PCA, from noisy examples.  Next, we focus on a significantly more challenging classification problem compared to the commonly used baselines in the previous work, namely classifying noisy versions of the Imagenet dataset. We show the utility of our approach on three different types of noise. The details of the hyperparameter tuning for different methods are given in Appendix~\ref{app:hyper} and the code for EG Reweighting  will be made available online.

\subsection{Noisy PCA}

In this experiment, we examine our EG Reweighting method on the PCA problem and show improvements when the examples are noisy.  Given a set of unlabeled training examples $\X = \{\x_i \in \R^d\}_{i=1}^{n}$, the goal of the PCA algorithm~\cite{pca} is to learn a subspace of dimension $k \ll d$ which minimizes the reconstruction loss of all examples. More formally, given a mean vector $\m \in \R^d$ and an orthonormal  matrix $\U \in \R^{d\times k}$ s.t. $\U^\top\U = \I_k$, the \emph{reconstruction loss} of the example $\x_i\in\X$ is defined as
\[
\ell(\m, \U|\,\x_i) = \Vert (\x_i - \m) - \U\U^\top (\x_i - \m)\Vert^2\, .
\]
The goal of the PCA algorithm is to find parameters $\{\m, \U\}$ that minimize the average reconstruction loss over all examples. Instead, we consider a reweighted version of the problem as in Eq.~\eqref{eq:cost_func_w} by maintaining a distribution $\w \in \Delta^{n-1}$ on all examples. The algorithm proceeds by alternatingly solving for $\{\m, \U\}$ and updating the weights $\w$ using EG. For a fixed $\w$, the parameter updates follow similar to~\cite{zhang2019robust}: the mean vector is replaced with a weighted average $\m = \sum_i w_i\, \x_i$ and $\U$ is set to the top-$k$ eigenvectors of the matrix $\sum_i w_i (\x_i-\m)(\x_i-\m)^\top$. 

To evaluate the performance of different algorithms, after computing the mean $\m$ and the subspace matrix $\U$, we report the average reconstruction loss on a set of clean test examples. More specifically, we split the datasets into training and test sets with a 90\%/10\% ratio, respectively and then add noise only to the training examples. We run experiments for three types of noise: 1)\emph{Gaussian random noise}, 2)\emph{occlusion noise}, and 3)\emph{Gaussian blur noise}. In occlusion noise, we corrupt $50\%$ of training images by placing a rectangular occlusion with a size chosen uniformly at random between $25$-$100\%$ the size of the image. The location of the occlusion is chosen randomly on the image and its pixels are randomly drawn from a uniform distribution.
In Gaussian random noise, we add a random Gaussian noise to the entire image where the noise std is randomly drawn from a beta distribution $\beta$($a=2,\,b = 5$). For blur noise, we convolve each image with a Gaussian filter with standard deviation drawn from the same beta distribution multiplied by $10$.

\begin{figure*}[t!]
\vspace{-0.2cm}
\begin{center}
    \subfigure[]{\includegraphics[width=0.45\linewidth]{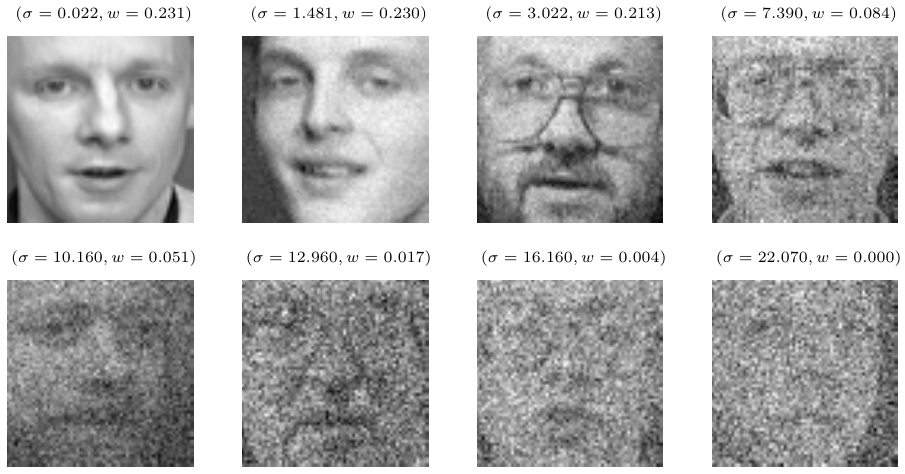}} \,\, \subfigure[]{\includegraphics[width=0.28\linewidth]{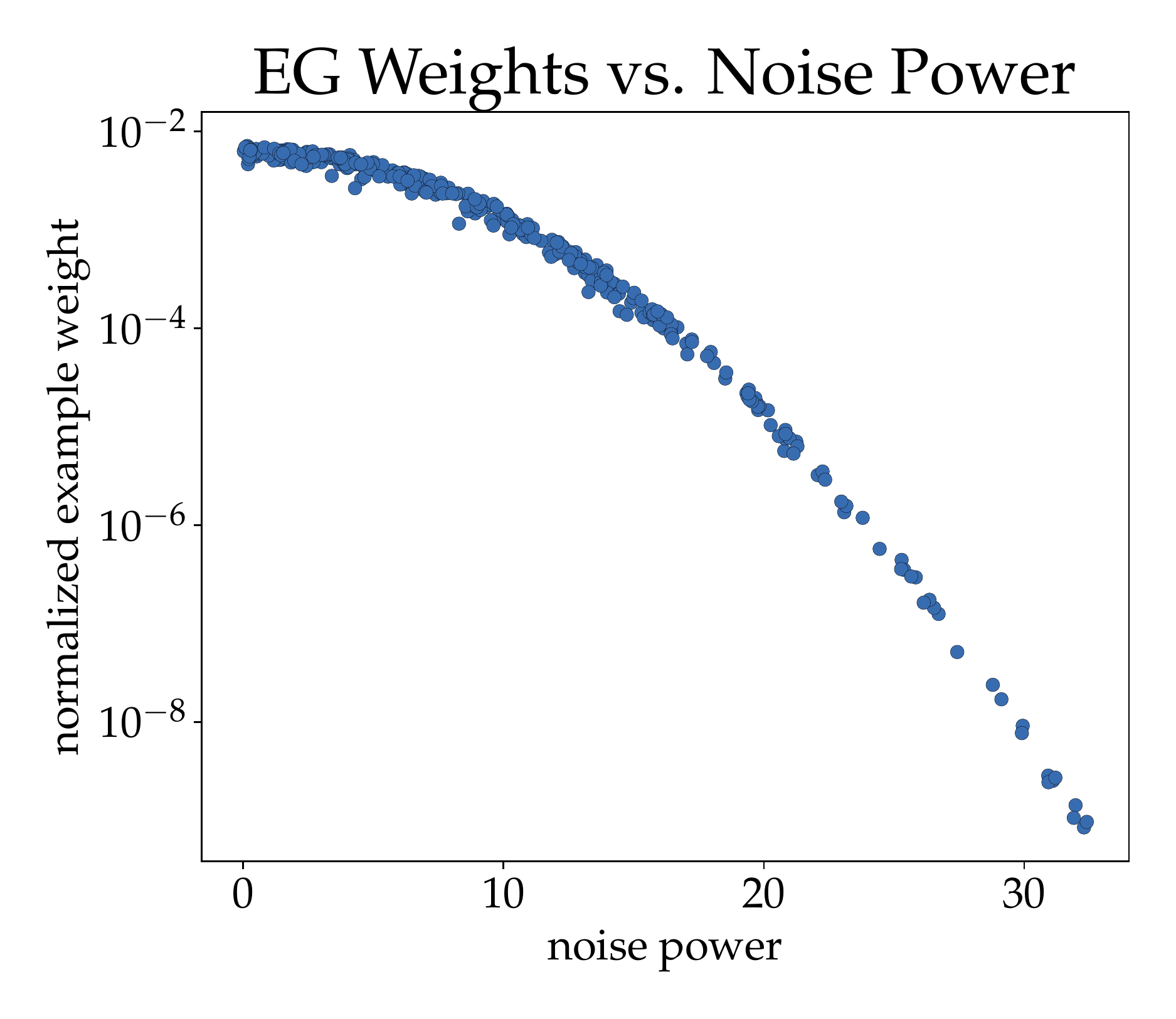}}
    \end{center}
    \vspace{-0.45cm}
    \caption{\textbf{Noisy PCA with EG Reweighting:} (a) A subset of noisy images from AT\&T dataset with increasing levels of  Gaussian noise power $\sigma$ (i.e. $\mathrm{L}_2$-norm of the additive noise component) along with their corresponding weights $w$ assigned by the EG Reweighting approach. (b) Weights of the examples vs. the noise power.}
    \label{fig:pca}
    \vspace{-0.3cm}
\end{figure*}

To evaluate the effect of our proposed regularization on EG, we also consider an alternative regularization based on capping~\cite{warmuth2008randomized} which upper-bounds the value of each weight by a constant value. Thus, we report the results for three versions of the EG Reweighting approach: (1) PCA with EG Reweighting (EGR-PCA), but no regularization, (2) PCA with capped EG (Capped EGR-PCA), and (3) PCA with regularized EG (Regularized EGR-PCA). We compare our results against vanilla PCA~\cite{pca}, robust 2DPCA with optimal mean (R2DPCA)~\cite{zhang2017auto}, capped robust 2DPCA with optimal mean (Capped R2DPCA)~\cite{zhang2017auto}, and robust PCA with RWL-AN~\cite{zhang2019robust}. We consider two benchmark face image datasets in our experiments, (i) AT\&T~\cite{ATT} which consists of $400$ images of size $64\times 64$ from $40$ classes, and UMIST~\cite{umist} which contains $575$ images of size $112\times 92$ from $20$ classes. Table~\ref{table:pca_compare} summarizes all of the experimental results. To account for the variance in results due to the random nature of the added noise, we run each algorithm $50$ times with different random seeds and reset the train-test at each run. It can be shown that for all three noises, EG, especially the regularized EGR-PCA, results in lower reconstruction losses on the test set compared to other algorithms. Fig.~\ref{fig:pca}(a) shows a subset of the noisy training examples from the AT\&T dataset that are corrupted with additive Gaussian noise. Each image is captioned by its noise power $\sigma$, i.e. norm of the additive component, and its weight $w$. It can be seen that as the examples become more corrupted, EG algorithm reduces their effect on the overall loss function by assigning smaller weights. Fig.~\ref{fig:pca}(b) shows the same result for all images in the dataset where we see that the weight of examples decreases as they become more noisy.

\begin{table}[t]
    \caption{\textbf{Results on the noisy PCA problems:} (Regularized) EG Reweighting outperforms the state-of-the-art robust PCA methods in all cases.\label{table:pca_compare}}
    \vspace{-0.2cm}
    \centering
    \resizebox{\columnwidth}{!}{
    \begin{tabular}{ l | c | c | c | c| c | c | c}
        \textbf{} & 
            \makecell{Vanilla \\ PCA} & 
            \makecell{RWL-AN} &
            \makecell{R2DPCA} & 
            \makecell{Capped \\ R2DPCA} &
            \makecell{EGR-PCA} &
            \makecell{Capped \\ EGR-PCA} &
            \makecell{Regularized \\ EGR-PCA}
            \\
        \hline
        AT\&T (random) & $37.48\pm1.37$ & $29.96\pm1.90$ & $30.67\pm0.22$ & $28.51\pm0.17$ & $28.47\pm1.71$ & $28.36\pm1.63$ & $\mathbf{25.92\pm1.42}$\\
        AT\&T (occlusion) & $23.06\pm0.19$ & $20.01\pm0.36$ & $28.85\pm0.06$ & $28.39\pm0.15$ & $19.85\pm0.28$ & $19.86\pm0.33$ & $\mathbf{19.82\pm0.36}$ \\
        AT\&T (blur) & $19.01 \pm 0.20$ & $19.41 \pm 0.32$ & $28.41 \pm 0.07$ & $28.40 \pm 0.08$ & $18.93 \pm 0.21$ & $18.99 \pm 0.15$ & $\mathbf{18.91\pm0.19}$\\
        \hline
        UMIST (random) & $100.48 \pm 2.94$ & $85.25 \pm 4.66$ & $115.47 \pm 0.45$ & $111.73 \pm 0.27$ & $79.63 \pm 3.75$ & $79.61 \pm 4.00$ & $\mathbf{73.77 \pm 3.61}$\\
        UMIST (occlusion) & $65.26 \pm 0.48$ & $58.75 \pm 1.36$ & $114.25 \pm 0.19$ & $114.22 \pm 0.23$ & $58.69 \pm 1.42$ & $58.54 \pm 1.41$ & $\mathbf{58.38 \pm 1.42}$\\
        UMIST (blur) & $55.97 \pm 0.57$ & $56.98 \pm 0.85$ & $110.99 \pm 0.11$ & $110.97 \pm 0.10$ & $56.02 \pm 0.50$ & $55.97 \pm 0.50$ & $\mathbf{55.78 \pm 0.50}$ \\
    \end{tabular}}
    \vspace{-0.5cm}
\end{table} 

\subsection{Noisy Imagenet Classification}

We consider classifying noisy versions of the Imagenet dataset~\cite{deng2009imagenet} which contains $\sim\!1.28$M examples from $1000$ classes. We use a ResNet-18 architecture~\cite{he2016deep} trained with SGD with heavy-ball momentum ($0.9$) optimizer with staircase learning rate decay 
and batch size of $1024$ for $240$ epochs. The baseline model achieves $72.58\pm0.09$\% top-1 accuracy on the test set. We consider three types of noise: 1) \emph{Symmetric label noise}: the labels of a random $40\%$ subset of the training examples are flipped, 2) \emph{JPEG compression noise}~\cite{poyser2020impact}: where the JPEG quality factors are sampled from a beta distribution $\beta$($a=0.9,\, b=0.8$) and multiplied by $100$. (a higher quality factor induces less noise.) The JPEG compression noise is highly non-uniform since for a fixed quality factor, the amount of perceivable noise largely depends on the content of the image~\cite{mier2021deep}. 3) \emph{Blur noise}: where we apply a Gaussian filter with  a kernel size of $5$ to each image (noise std $\sim\!\beta(a=1,b=0.2)$).

For comparison, we consider the following methods: 1) baseline trained with cross entropy loss, 2) baseline  trained with the bi-tempered cross entropy with softmax loss~\cite{bi-tempered} which provides a robust generalization of the standard cross entropy loss, 3) DCL~\cite{dataparams} which has similar complexity to our algorithm and has shown promising performance in settings with label noise, 4) our EG Reweighting method trained with cross entropy loss, 5) our EG Reweighting method combined with bi-tempered cross entropy loss. For each method, we tune the hyperparameters. For the combined EG additional Reweighting + bi-tempered loss, we do not perform any further tuning and simply combine the best performing hyperparameters for each case. Each result is averaged over $5$ runs.

For the label noise experiment, we apply the EG Reweighting algorithm on the training loss. For the JPEG compression and blur noise experiments, we rely on the negative of the \emph{visual quality score} from the NIMA model as the pseudo-loss. NIMA~\cite{talebi2018nima} is a deep convolutional neural network trained with images rated for aesthetic and technical perceptual quality. The baseline model used in NIMA is  Inception-v2~\cite{szegedy2016rethinking}. In~\cite{talebi2018nima}, the last layer of Inception-v2 is modified to 10 neurons to match the 10 human ratings bins in the AVA dataset~\cite{murray2012ava}. The original learning loss used in NIMA is the Earth Mover's Distance, yet, in the current framework we compute the mean score and use it in our hybrid loss. This approach is similar to the method used in~\cite{talebi2018learned} to employ NIMA as a training loss. As shown in \cite{talebi2018nima,talebi2018learned}, the NIMA loss is sensitive to image degradation such as blur, imbalanced exposure, compression artifacts and noise.

The results are shown in Table~\ref{table:imagenet}. EG Reweighting consistently outperforms the comparator methods in all cases. Additionally, combining EG Reweighting and the bi-tempered loss na\"ively provides further improvement in the label noise case. This shows the potential of combining the EG Reweighting method with a variety of loss functions. Also, we believe such improvements are possible for the JPEG compression and blur noise as well, however, this may require additional tuning of the hyperparameters. Note that methods such as DCL which reply on the training loss for handling noise yield even worse performance than the baseline in the case of JPEG compression and blur noise.

\begin{table}[t]
    \caption{\textbf{Results on the noisy Imagenet classification problems:} EG Reweighting consistently outperforms other methods. For the case of label noise, na\"ively combining EG Reweighting with bi-tempered loss using the best performing hyperparameters for each case improves the performance further.\label{table:imagenet}}\vspace{-0.3cm}
    \centering
    \resizebox{1\columnwidth}{!}{
    \begin{tabular}{c | c | c | c | c| c }
            \makecell{Noise type} & 
            \makecell{Baseline} &
            \makecell{\,Bi-tempered loss\,} & 
            \makecell{DCL} &
            \makecell{\,\,EG Reweighting\,\,} &
            \makecell{\,\,Bi-tempered loss\\ + \\EG Reweighting\,\,}
            \\
        \hline
        $40\%$ label noise & \,\,$65.36\pm 0.19$\,\, & \,\,$66.18 \pm 0.21$\,\, & \,\,$67.21\pm 0.09$\,\, & \,\,$67.44 \pm 0.11$\,\, & \,\,$\bm{67.63} \pm \bm{0.19}$\,\,\\
        \hline
        \,\,JPEG Compression noise\,\, & $70.88 \pm 0.18$ & $70.95 \pm 0.07$ & $70.63 \pm 0.16$ & $\bm{71.05} \pm \bm{0.02}$ & $70.91 \pm 0.20$\\
        \hline
        Blur noise & $71.01 \pm 0.16$ & $71.06 \pm 0.05$ & $70.83 \pm 0.10$ & $\bm{71.18} \pm \bm{0.11}$ & $70.87 \pm 0.14$
    \end{tabular}}
    \vspace{-0.3cm}
\end{table} 

\section{Conclusion}
The Exponentiated Gradient update \cite{eg} is one of the main updates developed mainly in the on-line learning context. It is based on trading off the loss with a relative entropy to the last weight vector instead of the squared Euclidean distance used for motivating Gradient Descent. A large number of techniques have been theoretically analyzed for enhancing the EG update such as specialist experts \cite{specialist}, 
capping the weights from above \cite{warmuth2008randomized} and lower bounding
the weights for handling shifting and sleeping experts \cite{trackexp,tracksmall}. We explore some of these techniques experimentally for the purpose of reweighting examples with the goal of increasing noise robustness. Surprisingly, these methods work well
experimentally for both supervised and unsupervised settings under a large variety of noise methods even though we introduce one additional weight per example. We found alternates to the specialist update and capping serving
the same function as the original that work
better experimentally. The modifications still clearly belong to the family of  relative entropy motivated updates. In future work, we are also planning to explore techniques for shifting and sleeping experts.

\bibliographystyle{plain}
\bibliography{refs}

\newpage
\appendix

\title{Exponentiated Gradient Reweighting for Robust Training Under Label Noise and Beyond\\
(Supplementary Material)}

\author{}
\institute{}
\maketitle

\section{Hyperparameter Tuning}
\label{app:hyper}

In this section, we provide more details about the hyperparameter tuning for the experiments.

\subsection{Noisy PCA Experiments}

In R2DPCA and capped R2DPCA the reduced dimensionalities are set to  $d_1 = 5$ and $d_2 = 5$, and in all other methods the reduced dimensionality is $k=d_1\times d_2=25$. In capped R2DPCA, for parameter $\epsilon$, we perform a grid-search on values $\{10,20,\ldots,50\}$. The integer parameter in RWL-AN is set to $r_s \times n$, where $r_s$ is the ratio parameter searched in a grid of $\{0.05,0.1,...,0.95\}$. As there is no warm-up phase for the PCA problem, we consider a learning rate of the form $\eta_0/t^\alpha$ where $\eta_0$ is the initial learning rate and $\alpha$ is tuned in the range  $\{0.6,0.65, ...,0.95\}$. The initial earning rate for EG is tuned in a grid of $\{10^{-2}, 10^{-1.5}, 10^{-1}, 10^{-0.5}, 1\}$. For capped EGR-PCA and regularized EGR-PCA, in addition to the learning rate and the decay factor $\alpha$, we tune the capping ratio in a set of $\{0.2,0.25,...,0.9\}$ and regularization factor in a set of $\{0.1,0.15,...,0.95\}$, respectively. 

The hyperparameters chosen for methods are as follows: 
\begin{enumerate}
    \item For RWL-AN, $r_s=0.65$ for UMIST(blur), $r_s=0.6$ for UMIST(random), $r_s=0.8$ for UMIST(occlusion), $r_s=0.25$ for AT\&T(blur), $r_s=0.95$ for AT\&T(occlusion), and $r_s=0.45$ for AT\&T(random)
    
    \item For capped R2DPCA, $\epsilon=10$ for all kinds of noise on AT\&T, $\epsilon=50$ for UMIST(occlusion), $\epsilon=20$ for UMIST(blur) and UMIST(random)
    
    \item For EGR-PCA, $(\eta_0, \alpha) = (0.01, 0.7)$ for UMIST(blur), $(\eta_0, \alpha) = (0.01, 0.95)$ for UMIST(random), $(\eta_0, \alpha) = (0.01, 0.6)$ for UMIST(occlusion), $(\eta_0, \alpha) = (0.01, 0.6)$ for AT\&T(blur), $(\eta_0, \alpha) = (0.032, 0.8)$ for AT\&T(random), $(\eta_0, \alpha)$ $= (0.032, 0.6)$ for AT\&T(occlusion).
    
    \item For Capped EGR-PCA, $(\text{capping-ratio}, \eta_0, \alpha) = (0.25, 0.01, 0.75)$ for UMIST (blur), $(\text{capping-ratio}, \eta_0, \alpha) = (0.45, 0.32, 0.75)$ for UMIST(random), $(\text{capp-}$\\
    $\text{ing-ratio}, \eta_0, \alpha) = (0.45, 0.01, 0.8)$ for UMIST(occlusion), $(\text{capping-ratio},$\\ $\eta_0, \alpha) = (0.45, 0.01, 0.6)$ for AT\&T(blur), $(\text{capping-ratio},\eta_0, \alpha) = (0.35, 0.1,$\\ $ 0.85)$ for AT\&T(random), $(\text{capping-ratio}, \eta_0, \alpha) = (0.4, 0.032, 0.6)$ for AT\&T\\(occlusion). 
    
    \item For regularized EGR-PCA,  $(\text{reg-factor}, \eta_0, \alpha) = (0.15, 0.1, 0.68)$ for UMIST\\(blur),  $(\text{reg-factor}, \eta_0, \alpha) = (0.5, 0.1, 0.95)$ for UMIST(random),  $(\text{reg-factor}, \eta_0,$\\$ \alpha) = (0.35, 0.032, 0.65)$ for UMIST(occlusion),  $(\text{reg-factor}, \eta_0, \alpha) = (0.5, 0.32,$\\$ 0.9)$ for AT\&T(blur),  $(\text{reg-factor}, \eta_0, \alpha) = (0.45, 0.32, 0.95)$\\ for AT\&T(random),  $(\text{reg-factor}, \eta_0, \alpha) = (0.35, 0.1, 0.6)$ for AT\&T(occlusion).
\end{enumerate}

\subsection{Noisy Imagenet Experiments}

In noisy Imagenet experiments we tune hyperparameters in the suggested range for all methods: for bi-tempered loss, we tune $t_1$ and $t_2$ in the range $[0.8, 0.99]$, $[1.05, 1.2]$, respectively. For DCL, we set set the regularizer to $5e$-$4$ and tune the scale learning rate for the examples in the range $[0.01, 1.0]$ and set the momentum parameter to $0.9$. We tune the learning rate and the regularizer factor $r$ as in the previous section and use a linear ramp-up until $20$ epochs followed by a decay of $0.9$ every $30$ epochs. For the combined EG additional Reweighting + bi-tempered loss, we do not perform any tuning and simply combine the best performing hyperparameters for each case. The list of hyperparameters for different methods are as follows: (1) Bi-tempered loss: $(t_1, t_2) = (0.92, 1.1)$ for label noise and $(t_1, t_2) = (0.95, 1.05)$ for JPEG compression and blur noise experiments. We start all temperatures from $1.0$ (i.e. cross entropy with softmax) and linearly interpolate to the final value in the first over the first $10$ epochs. (2) DCL: we set the example scale learning rate to $0.05$ in all experiments (all the non-trivial values of the learning rate deteriorated the performance compared to the baseline for the JPEG compression and blur noise). We do not apply any scaling based on the class. (3) EG Reweighting: we set the learning rate and the regularizer to $0.1$ and $0.9$, respectively, for label noise, $0.02$ and $1.0$ for JPEG compression noise, and $0.05$ and $1.0$ for blur noise. 

\end{document}